\documentclass[runningheads]{llncs}
\usepackage{amsmath}
\usepackage{algorithm}
\usepackage{algpseudocode}
\usepackage{xcolor}
\usepackage{colortbl}
\usepackage{multirow}
\usepackage{multicol}
\usepackage{graphicx}
\usepackage{booktabs}
\usepackage{amsfonts}
\begin{document}
\title{Towards Safer Mobile Agents: Scalable Generation and Evaluation of Diverse Scenarios for VLMs}
%
\author{Takara Taniguchi \and
Kuniaki Saito \and
Atsushi Hashimoto}
\authorrunning{T. Taniguchi et al.}
%
\institute{OMRON SINIC X,
Nagase Hongo Building 3F 5-24-5 Hongo, \\ Bunkyo-ku Tokyo-to, Japan
}
\maketitle              

\begin{abstract}
Vision Language Models (VLMs) are increasingly deployed in autonomous vehicles and mobile systems, making it crucial to evaluate their ability to support safer decision-making in complex environments. 
However, existing benchmarks inadequately cover diverse hazardous situations, especially anomalous scenarios with spatio-temporal dynamics. 
While image editing models are a promising means to synthesize such hazards, it remains challenging to generate well-formulated scenarios that include moving, intrusive, and distant objects frequently observed in the real world. 
To address this gap, we introduce \textbf{HazardForge}, a scalable pipeline that leverages image editing models to generate these scenarios with layout decision algorithms, and validation modules. 
Using HazardForge, we construct \textbf{MovSafeBench}, a multiple-choice question (MCQ) benchmark comprising 7,254 images and corresponding QA pairs across 13 object categories, covering both normal and anomalous objects.  
Experiments using MovSafeBench show that VLM performance degrades notably under conditions including anomalous objects, with the largest drop in scenarios requiring nuanced motion understanding.
\end{abstract}    

\section{Introduction}\label{sec:intro}
\begin{figure}[t]
    \centering
    \includegraphics[width=1.0\linewidth]{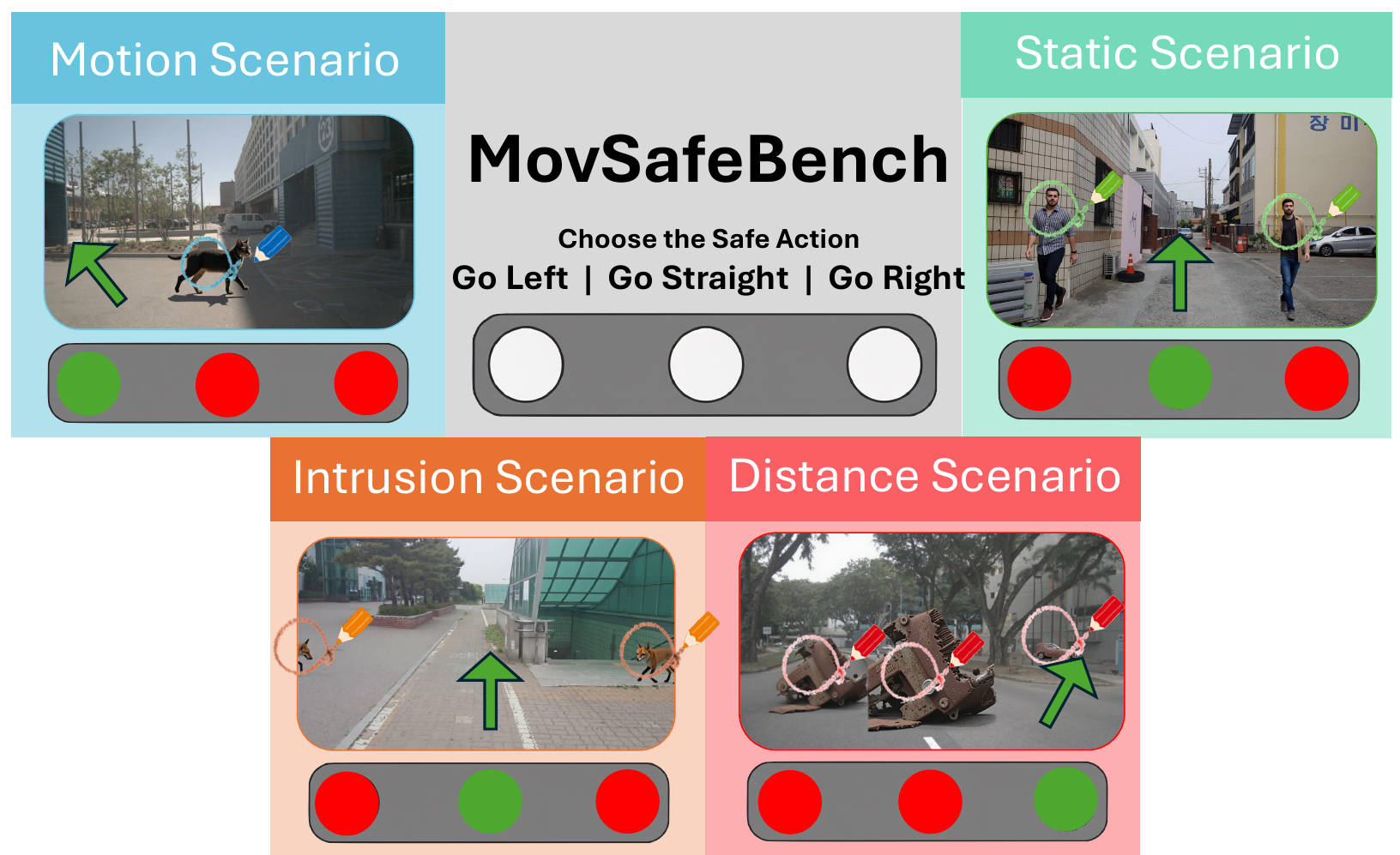}
    \caption{Overview of MovSafeBench.
    Our benchmark assesses how reliably Vision-Language Models (VLMs) operate and how well they visually ground information when used in mobile agents, including vehicles and mobile robots.
    }
    \label{fig:intro_overview}
\end{figure}

Vision Language Models (VLMs)~\cite{guo2025surds,sima2024drivelm,marcu2024lingoqavisualquestionanswering,tian2024drivevlmconvergenceautonomousdriving,xie2025DriveBenchvlmsreadyautonomousdriving,shi2025scvlmenhancingvisionlanguagemodel,li2025vladbenchfinegrainedevaluationlargevisionlanguage,wang2025metavqa} are increasingly integrated into autonomous mobile agents, such as self-driving vehicles~\cite{xie2025DriveBenchvlmsreadyautonomousdriving,ye2025safedriverag} and mobile robots~\cite{savlm_icra2025,payandeh2024socialllavaenhancingrobotnavigation}.
As these agents operate in open, real-world settings and interact directly with people and infrastructure, ensuring the safety of VLM-driven situational judgment is paramount.

Evaluating the safety of VLMs in determining movement actions for mobile platforms, including autonomous vehicles and mobile robots, is important as safe decision-making is addressed for effective collision avoidance in previous research.
In autonomous driving, safety is typically operationalized as adherence to traffic regulations and physical constraints, together with the avoidance of collisions~\cite{xie2025DriveBenchvlmsreadyautonomousdriving,wei2025driveqapassingdrivingknowledge}. 
For mobile robots, safety considerations emphasize the reduction of environmental risk by avoiding collisions with people and obstacles and by preventing falls or tip-overs~\cite{savlm_icra2025,payandeh2024socialllavaenhancingrobotnavigation}.

The scarcity of data that captures common and anomalous scenarios, including their temporal dynamics and spatial relations, limits the development of benchmarks for safer decision-making.
There are two scenes encountered by mobile agents, classified into two categories: common scenarios and anomalous scenarios.
Common scenarios involve ordinary objects in high-risk contexts, such as vehicles on the road or pedestrians crossing a lane, whereas anomalous scenarios correspond to rare events involving unusual objects, such as deer or pigs appearing on the roadway. 
In both cases, temporal dynamics and spatial relations are crucial for evaluating VLMs’ ability to reason about object motion and distance, respectively.
Acquiring such dynamic scenes is challenging, as both common and anomalous scenarios occur only rarely in real-world data.

While image editing models~\cite{wu2025qwenimagetechnicalreport,labs2025flux1kontextflowmatching} are commonly used to synthesize such dynamic scenarios, generating these scenarios remains difficult using only text prompts.
For comparable evaluation, scenario generation should be represented as predefined templates specifying object placement and orientation, constraining safety-critical interactions such as motion, intrusion, and distance conditions involving moving objects, intrusive objects, and far-field objects.
However, image editing models can fail to generate these objects when relying solely on text prompts due to limited prompt alignment ability. 

To address these challenges, we propose \textbf{HazardForge}, a scalable pipeline that addresses (i) data scarcity by scalable synthesis of both common and anomalous hazards, and (ii) spatio-temporally diverse generation.
HazardForge employs image editing models~\cite{wu2025qwenimagetechnicalreport,labs2025flux1kontextflowmatching} to generate diverse types of objects across common and anomalous ones.
Additionally, we formulate four scenarios to evaluate the safe decision-making capability of VLMs in spatio-temporally diverse situations for comparable evaluation.
Fig.~\ref{fig:intro_overview} shows four formulated scenarios that capture diverse real-world situations by featuring a moving hazard, an off-screen intrusion into the ego lane, and a far-field hazard object, alongside a default static setting, thereby probing motion understanding, intrusion understanding, distance understanding, and default static scene understanding of VLMs, respectively.
To generate the four formulated scenarios, HazardForge incorporates algorithms for conditional mask layout selection for each scenario and VLM-based validity checking, which verifies whether the generated objects match the prompt-specified attributes, such as placement and orientation.

Using HazardForge, we construct \textbf{MovSafeBench}, a benchmark that evaluates safety across mobile agents, including both autonomous vehicles and mobile robots.
MovSafeBench is a benchmark composed of 7,254 images, with four scenarios and 13 types of objects, ranging from general objects to anomalous objects.

We evaluate seven VLMs on MovSafeBench to compare their ability for safe decision making, considering the spatio-temporal information with both common and anomalous objects.
Experimental results show the limited ability of VLMs to decide a safe action in a scenario with anomalous objects compared to common objects. 
Scenario-wise analysis further shows that VLMs struggle to understand the motion of both common and anomalous objects.

\section{Related work}\label{sec:related}








\begin{table}[htbp]
\caption{Comparisons among evaluation mobility benchmarks, including mobile robots and autonomous vehicles.
}\label{tbl:related}
\centering
\setlength{\tabcolsep}{18pt}
\resizebox{\textwidth}{!}{
\begin{tabular}{@{}crrcccc@{}}
\toprule
Benchmarks    & \# Images & \# QA pairs & Robots & Vehicles & Anomalous obj. & Metrics \\ \midrule
DriveLM~\cite{sima2024drivelm}      & 4,794        & 15,480    & $\times$      & $\surd$            & $\times$         & Language, GPT      \\
DriveBench~\cite{xie2025DriveBenchvlmsreadyautonomousdriving}    & 19,200       & 20,498    & $\times$      & $\surd$            & $\times$         & Acc, Language, GPT \\
DriveQA-V~\cite{wei2025driveqapassingdrivingknowledge}     & 68,000       & 448,000   & $\times$      & $\surd$            & $\times$         & Acc, Language      \\
SA-Bench~\cite{savlm_icra2025}      & 1,000        & 1,000     & $\surd$       & $\times$           & $\times$         & Language, GPT      \\
SNEI~\cite{payandeh2024socialllavaenhancingrobotnavigation}          & 1,961        & 1,961     & $\surd$       & $\times$           & $\times$         & GPT                \\
SocialNav-SUB~\cite{munje2025socialnavsubbenchmarkingvlmsscene} & \multicolumn{1}{c}{-}           & 4,968     & $\surd$       & $\times$           & $\times$         & Accuracy           \\
MovSafeBench (Ours)  & 7,254        & 7,254     & $\surd$       & $\surd$            & $\surd$          & Accuracy           \\ \bottomrule
\end{tabular}
}
\end{table}

\subsection{VLMs for Mobility}
Vision language models (VLMs)~\cite{bai2025qwen25vltechnicalreport,NEURIPS2023_llava,wang2025internvl35advancingopensourcemultimodal,microsoft2025phi4minitechnicalreportcompact,steiner2024paligemma2familyversatile} have demonstrated strong cross-modal reasoning across diverse environments and are increasingly applied to mobility domains, notably in autonomous driving~\cite{guo2025surds,sima2024drivelm,marcu2024lingoqavisualquestionanswering,tian2024drivevlmconvergenceautonomousdriving,xie2025DriveBenchvlmsreadyautonomousdriving,shi2025scvlmenhancingvisionlanguagemodel,li2025vladbenchfinegrainedevaluationlargevisionlanguage,wang2025metavqa} and social navigation for mobile robots~\cite{savlm_icra2025,payandeh2024socialllavaenhancingrobotnavigation,munje2025socialnavsubbenchmarkingvlmsscene}. 
Safety has therefore emerged as a central concern in VLM-enabled mobile agents: in autonomous driving, methods such as SafeDriveRAG~\cite{ye2025safedriverag} aim to bolster performance under safety-critical conditions, while in robotics, SA-VLMs~\cite{savlm_icra2025} and Social-LLaVA~\cite{payandeh2024socialllavaenhancingrobotnavigation} target safe behavior and collision avoidance in social environments. 
A key bottleneck across these domains is the scarcity of long-tail anomalous data, partly because inherently dangerous scenarios are rare and difficult to collect. 
To close this gap, our approach automatically generates both common and anomalous scenarios, with objects that vary over time and are diverse in their spatial configurations, enabling systematic evaluation of VLM robustness and safety in real-world mobility settings.

\subsection{Benchmark and Datasets}
Table~\ref{tbl:related} shows existing benchmarks for mobile robots and autonomous vehicles, and the position of our work.
In autonomous driving, language-based datasets and benchmarks including open-ended VQA and MCQ have been proposed~\cite{marcu2024lingoqavisualquestionanswering,kim2018bddx,deruyttere-etal-2019-talk2car,sacheva2024rank2tell,Qian_2024_nusceneQA_AAAI,malla2023dramajointrisklocalization,wu2025nuPromptlanguage,tian2024drivevlmconvergenceautonomousdriving,guo2025surds}.
These datasets and benchmarks are constructed from established large datasets~\cite{sun2020WAYMOopenscalabilityperceptionautonomousdriving,Caesar_2020_CVPR_nuscenes,2023kitti,kim2018bddx} or originally collected data in the real world with language annotations.
Simulation-based methods, frequently employing CARLA~\cite{pmlr-v78-dosovitskiy17aCARLA}, are a feature of some works~\cite{wang2025metavqa,sima2024drivelm}.
These benchmarks handle diverse topics, including the ability of a steering decision considering spatial relation~\cite{sima2024drivelm}, robustness for weather conditions~\cite{xie2025DriveBenchvlmsreadyautonomousdriving}, and compliance of traffic regulations~\cite{wei2025driveqapassingdrivingknowledge,ye2025safedriverag,zhang2025SCDbenchevaluationsafetycognitioncapability}.
In mobile robots, language-based datasets are also proposed~\cite{payandeh2024socialllavaenhancingrobotnavigation,savlm_icra2025,munje2025socialnavsubbenchmarkingvlmsscene}, which are also constructed from large datasets~\cite{karnan2022SCANDsociallycompliantnavigationdataset,park2020sideguide}.
The target of these benchmarks on mobile robots is to measure the ability of safer decision-making.
The problem with the existing benchmark is the lack of hazardous scenarios of real-world data, including anomalous objects with spatio-temporal diversity.
In our study, we focus on creating a benchmark for safer decision-making with our pipeline generating anomalous objects in the real-world scenes.

\section{HazardForge}
\label{sec:HazardForge}
We introduce HazardForge, a pipeline designed to automatically generate spatio-temporally diverse scenarios that include both common and anomalous objects, enabling evaluation of mobile agents' long-tail dangerous situations. 
The main role of HazardForge is generating oriented, intrusive, and distant objects in a formulated manner with the verification module, addressing limitations inherent in approaches that rely solely on text-based prompts due to less aligned generation.

Given a triplet of image, question, and ground-truth action indicating the direction of the safe direction for mobile agents, our goal is to render anomalous objects in the image to simulate long-tail dangerous situations, while ensuring that the ground-truth action remains unchanged.  
To comprehensively model such long-tail scenarios, we diversify (i) the object categories, and (ii) the spatial locations and the moving direction of the inserted objects within the scene while considering the consistency with the ground-truth action. 

In Sec.~\ref{sec:preliminary}, we describe the preliminary for our task definition, categories of rendered objects, and an image editing module. 
In Sec.~\ref{subsec:pipeline}, we describe the overall pipeline to render the objects considering the diversity of their spatial locations and the moving directions. 

\subsection{Preliminary}\label{sec:preliminary}

\noindent\textbf{Direction-oriented question answering}
In our study, we evaluate the capability of VLMs to identify the safe direction for mobile agents in an anomalous scene created by our pipeline. 
For ease of evaluation, we frame the task as direction-oriented question answering, where the model selects the safe driving direction (right, left, or straight) given an input image and question. Specifically, given an image ($I$) and a question ($q$), the model is asked to select the correct action $a$ from three options: $a_\text{R}$ (“go right”), $a_\text{L}$ (“go left”), and $a_\text{C}$ (“go straight”). 
In the following, we denote the ground-truth action as $a_{k^*}$ where $k^* \in \{\text{L},\text{C},\text{R} \}$.
We assume access to the dataset with this formulation and aim to create an anomalous scene based on the image and ground-truth safe direction. 

\noindent\textbf{Rendered categories}
We aim to construct a benchmark that evaluates the performance of VLMs in recognizing both common and uncommon objects encountered by diverse mobile agents. Specifically, we select four common object categories—\textit{human, motorcycle, bicycle, and cone}—that frequently appear on roads, and nine anomalous object categories—\textit{rocks, debris, roadkill, dog, cat, deer, fox, pig, and raccoon}—that are rarely observed in standard mobile-agent datasets. 

\noindent\textbf{Image editing module}
Given an image $I_{\text{in}}$ and a conditional mask $m \in \Omega(I_{\text{in}})$, and a prompt $t \in \mathcal{T}$, the output image $I_{\text{out}} \in \mathcal{I}$ is generated by sampling from the model as follows:
\begin{align*}
I_{\text{out}} \sim M(I_{\text{in}}, m, t),
\end{align*}
The text prompt defines both the object’s category and its direction of motion.
The direction of motion is selected from the options \textit{facing right}, \textit{facing left}, \textit{facing forward}, and \textit{facing backward}.
Accordingly, the prompt $t$ is written in a simple form such as \textit{“Render a \{object category\}, \{direction\}.”}
The mask $m$ determines where the object is positioned in the scene.

\noindent\textbf{Image regions to render objects}
Given the three possible directions that the agent can choose, we horizontally split the image into three regions to clarify the location to render objects. Assuming an image $I \in \mathbb{R}^{H \times W \times 3}$, we define an image region $\Omega(I) = \{(x,y) \mid 0 \le x < W,\; 0 \le y < H\}$ whose origin is at the bottom-left corner.
Each action in $\{a_\text{L},a_\text{C},a_\text{R} \}$ corresponds to the destination defined as $
\Omega_\text{L}(I) = \{(x,y) \in \Omega(I) \mid 0 \leq x < \tfrac{W}{3}\}, 
\Omega_\text{C}(I) = \{(x,y) \in \Omega(I) \mid \tfrac{W}{3} \leq x < \tfrac{2W}{3}\}, 
\Omega_\text{R}(I) = \{(x,y) \in \Omega(I) \mid \tfrac{2W}{3} \leq x < W\}$, respectively. 
Depending on the ground-truth action and the variation of the spatial locations and moving directions of objects to insert, our pipeline chooses the split region as the target location for object rendering.

\subsection{Pipeline}\label{subsec:pipeline}

\begin{figure}
    \centering
    \includegraphics[width=1\linewidth]{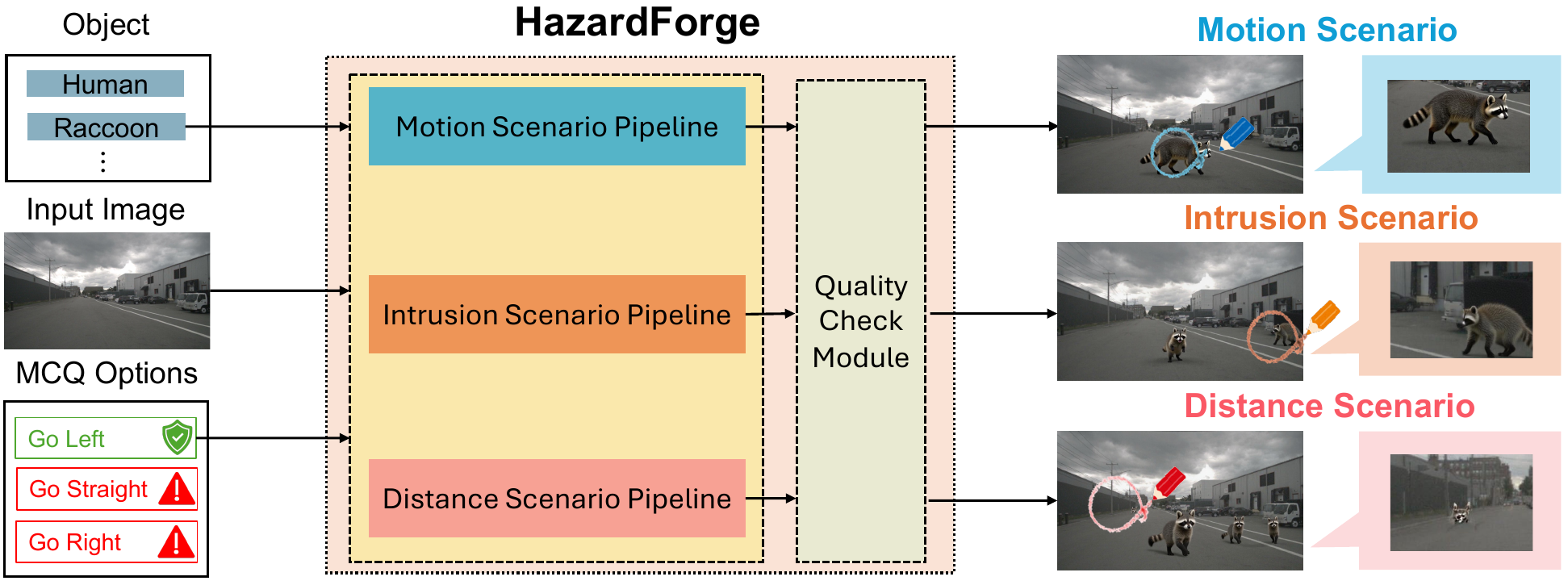}
    \caption{Overall workflow of our HazardForge. HazardForge generates the scene, including moving objects, intrusive objects, and remote objects.}
    \label{fig:scenario_overview}
\end{figure}

\subsubsection{Overview}
As shown in Fig.~\ref{fig:scenario_overview}, our pipeline diversifies the input image by the product of 13 categories and three generation scenarios consisting of \textit{motion scenario}, \textit{intrusion scenario}, and \textit{distance scenario} according to the ground-truth of the question. 

\subsubsection{Motion scenario}\label{subsec:motion}
When an object moves in a certain direction, maneuvering around it is often the safest decision. 
Grasping how the object moves is essential for ensuring that the VLM can safely maneuver it.
The motivation of the \textit{motion scenario} is to evaluate VLMs' ability to understand motion direction and make safe decisions to go around the object.

To simulate the scenes from an input image $I_{\text{in}} \in \mathcal{I}$ for maneuvering, we define the scenario where (i) the ground-truth action is restricted to either $a_\text{R}$ or $a_\text{L}$, and (ii) an object is positioned at the center region $\Omega_\text{C}(I_{\text{in}})$, moving toward the side opposite to the ground-truth region.
The object's motion direction specified by the prompt $t$ is set to \textit{facing left} when the ground-truth action is $a_\text{R}$, and \textit{facing right} when the ground-truth action is $a_\text{L}$ to make the ground-truth region appear safer than the other image regions.
We adopt the mask $\Omega_\text{C}(I_{\text{in}})$ as input, because the object must be synthesized at the center of the image.
Then, the image editing model produces an output image $I_{\text{out}} \sim M(I_{\text{in}}, \Omega_\text{C}(I_{\text{in}}), t)$, where the object is generated at the center of the image and oriented according to the direction specified by the prompt $t$.
\subsubsection{Intrusion scenario}\label{subsec:intrusion}
\begin{figure}[htbp]
    \centering
    \includegraphics[width=1.0\linewidth]{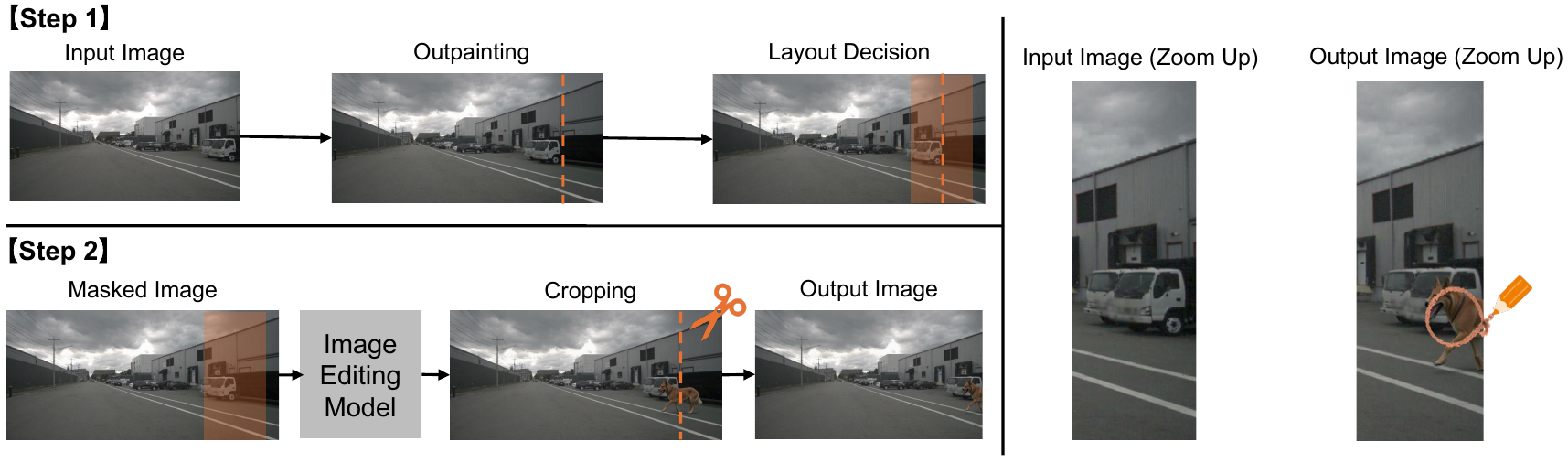}
    \caption{The detailed workflow of our \textit{intrusion image editing}.}
    \label{fig:padding}
\end{figure}
In real-world environments, intrusions from outside the field of view occur frequently, posing significant safety concerns for VLMs integrated with mobile agents. 
The purpose of \textit{intrusion scenario} is to generate such scenes from the input image $I_\text{in} \in \mathcal{I}$ to evaluate the VLM's capability to comprehend intrusion events.

In \textit{intrusion scenario}, the object is placed other than the ground-truth region $\Omega_{k^*}(I_\text{in})$, with intrusion effects modeled at the image’s side boundary to make the ground-truth region appear comparatively safer than the alternatives.
A nearby object is generated in the central image region $\Omega_{\text{C}}(I_\text{in})$ with a front-facing view\footnote{The generation method of front-facing view is detailed in the Appendix.}. 
At the boundary of the region in $\Omega_{\text{R}}(I_\text{in})$ or $\Omega_{\text{L}}(I_\text{in})$, the object is generated to simulate an intrusion.
However, generating a partially visible object is difficult with only text prompts and masks in a traditional way.

To create these intrusive objects, we introduce \textit{intrusion image editing}, which generates an object emerging from outside the field of view.
We utilize \textit{intrusion image editing} because generating a partially visible object using only text prompts and masks in the conventional approach is challenging.
Fig.~\ref{fig:padding} shows the workflow of \textit{intrusion image editing} to generate objects that appear from the outside of the sight.
Step 1 in Fig.~\ref{fig:padding} illustrates the workflow of the layout decision used to create an object at the boundary of the image.
We convert the original image $I_{\text{in}} \in \mathbb{R}^{H \times W \times 3}$ into an $r \in \mathbb{R}$ pixel outpainted image $I_{\text{in}}' \in \mathbb{R}^{H \times (W + r) \times 3}$, enabling the subsequent natural insertion of the object into the image for Step 2.
Denoting the parameter $l \in \mathbb{R}$ to adjust the size of the mask, the configuration of the conditional mask to create the object positioned at the boundary of the scene is defined as follows.
\begin{align*}
m_{\text{pad}} =
\begin{cases}
[r -l, r + l) \times [0, H), & \text{if } a = a_\text{L}, \\
[W - r - l, W - r + l) \times [0, H), & \text{if } a = a_\text{R},
\end{cases}   
\end{align*}
Given a prompt $t \in \mathcal{T}$ specifying the object category and its orientation, we produce the object-integrated image $I_{\text{pad}} \sim M(I_{\text{in}}', m_{\text{pad}}, t)$ as illustrated in Step 2 of Fig.~\ref{fig:padding}. 
By cropping $I_{\text{pad}} \in \mathbb{R}^{H \times (W + r) \times 3}$ to remove $r$ pixels from the border, thereby forming a naturally intrusive object, we derive the final output image $I_{\text{out}} \in \mathbb{R}^{H \times W \times 3}$.

\subsubsection{Distance scenario}\label{subsec:distance}
\begin{figure}[htbp]
    \centering
    \includegraphics[width=1.0\linewidth]{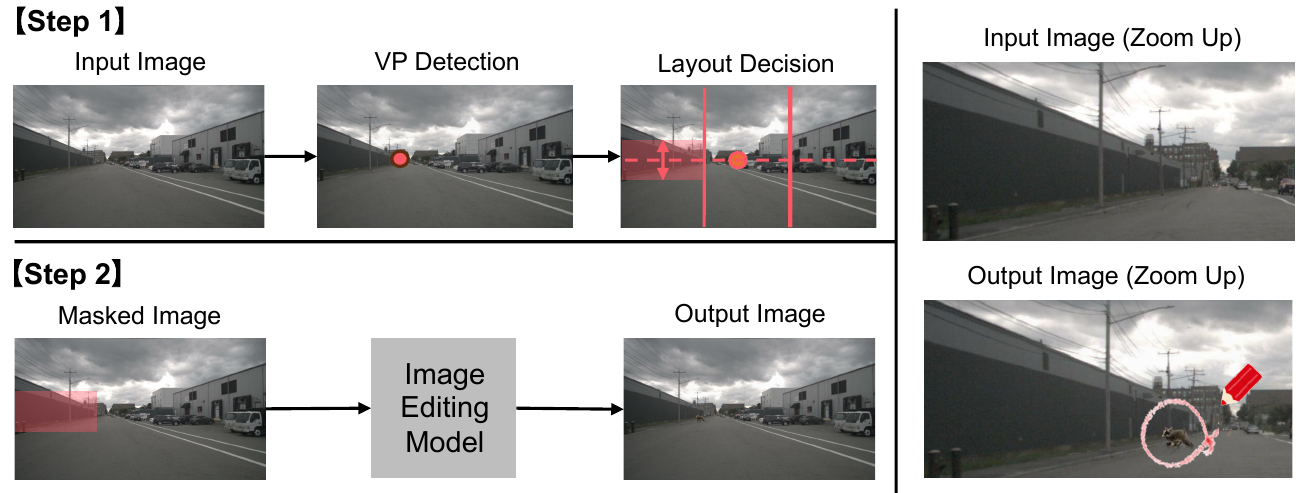}
    \caption{The detailed workflow of our \textit{distance image editing}.}
    \label{fig:vp}
\end{figure}
For mobile agents, nearby objects are more critical than distant ones, because a nearby object is more likely to cause a collision than a remote object.
The motivation of \textit{distance scenario} is to evaluate whether VLMs correctly interpret the nearby object as unsafe rather than focusing on the distant one.

\textit{Distance scenario} includes a remote object in the ground-truth image region and nearby objects in the other image region generated from an input image $I_\text{in} \in \mathcal{I}$ to show that the ground truth region is safer than the others.
While we perform the image editing model to generate the nearby object with a front-facing view into all regions except the ground-truth region $\Omega_{k^*}(I_\text{in})$, a distant object is also generated at the ground-truth region $\Omega_{k^*}(I_\text{in})$.
However, generating an object that appears at a specified distance from the ego mobile agent is difficult with only text prompts for traditional image editing models.

To overcome this challenge, we introduce \textit{distance image editing}, which creates an object that appears to be located far away within the scene.
Fig.~\ref{fig:vp} shows the overall workflow of our \textit{distance image editing}.
Step 1 of the Fig.~\ref{fig:vp} shows the layout decision procedure, which determines the area of the object to be generated in the ground-truth region $\Omega_{k^*}(I_\text{in})$.
Vanishing point (VP) detection is utilized to determine the layout of the conditional mask where the remote object is to be generated, since identifying the VP allows us to approximate where objects should be generated to appear further away.
In the input image $I_{\text{in}}$, the VP is represented as $p = (x',y') \in \Omega(I_\text{in})$ and is determined using the rule-based approach described in~\cite{luvpdetect2017}.
Denoting the parameter $d \in [0,1]$ to adjust the size of the mask, the layout of the conditional mask in the safer region $\Omega_{k^*}(I_\text{in})$ is determined to generates the remote object in $\Omega_{k^*} (I_\text{in})$ as follows.
\begin{align*}
m_{k^*}^{\text{dist}} = \{ (x,y) \in \Omega_{k^*}(I_\text{in}) \mid |y - y'| \le d \cdot H \}.
\end{align*}
Given in Step 2 of Fig.~\ref{fig:vp}, we obtain the final sampling result $I_{\text{out}} \sim M(I_{\text{in}}, m_{k^*}^{\text{dist}}, \text{Concat}(t,t_\text{dist}))$, where the pre-defined prompt $\text{Concat}(t,t_\text{dist})$ indicates the smaller object such as \textit{$t$, make the object smaller”} to make the object appear far away.

To compare the effect of each scenario, we additionally create the \textit{static scenario}, where the nearby objects with the front view are generated other than the ground-truth region $\Omega_{k^*}(I_\text{in})$.
The specifics of the \textit{static scenario} are described in the Appendix.

\subsubsection{Quality check}\label{subsec:vlmcheck}
\begin{figure}[htbp]
    \centering
    \includegraphics[width=1.0\linewidth]{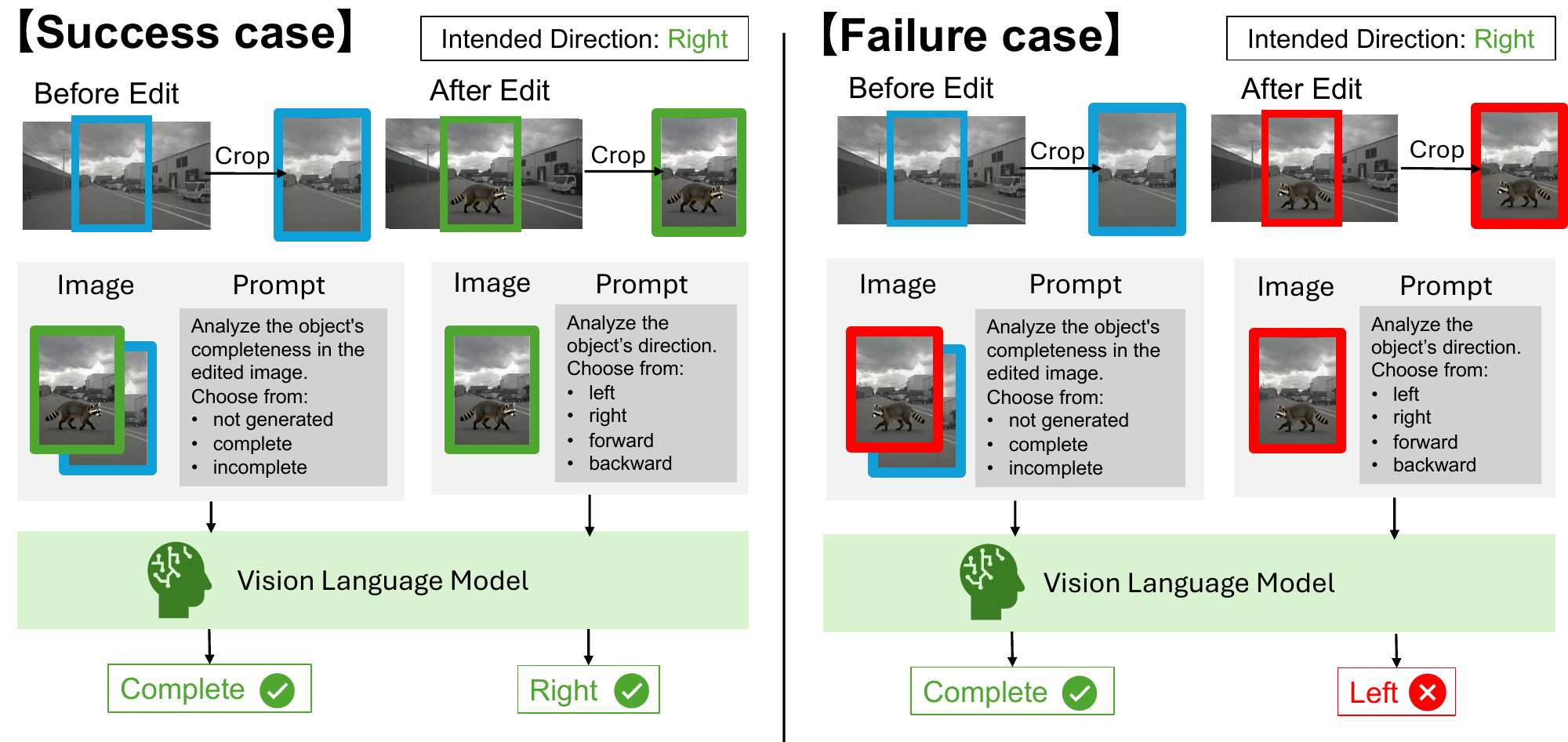}
    \caption{The detailed workflow of our quality check module.}
    \label{fig:vlmcheck}
\end{figure}
The successful creation of high-quality datasets hinges upon ensuring that the generation process in each scenario proceeds as intended. 
However, image-editing models occasionally exhibit behavior that fails to produce any objects or deviates from the desired direction of generation.
The purpose of quality checking is to verify that the generated objects are complete and correctly oriented to filter out these unintended generations.

We employ a VLM-based quality check module that independently verifies (i) whether the object is fully inserted and (ii) whether it moves in the intended direction specified by the prompt $t \in \mathcal{T}$.
We utilize Qwen3-VL-30B~\cite{bai2025qwen3vltechnicalreport} for VLM of our quality check module.
Fig.\ref{fig:vlmcheck} shows the overall workflow of the quality check module.
We execute cropping for each object mask to only compare the region where the object is generated.
For each output image $I_{\text{out}} \sim M(I_{\text{in}}', m, t)$ where the input image $I_{\text{in}} \in \mathcal{I}$, mask $m$, and a prompt $t \in \mathcal{T}$, we crop the region of $I_{\text{in}}$ and $I_{\text{out}}$ covered by the mask $m$.
We denote the cropped regions of $I_{\text{in}}$ and $I_{\text{out}}$ as $I_{\text{in}}'$ and $I_{\text{out}}'$, respectively.
The VLM first verifies whether the object is completely generated in the designated region specified by the mask $m$.
Cropped images $I_{\text{in}}'$, $I_{\text{out}}'$, and a prompt such as \textit{“Analyze the object's completeness in the edited image. Choose from: not generated, complete, incomplete”} are produced as inputs to the VLM.  
The result is accepted only if the VLM's output is \textit{complete}. 
In addition, the VLM checks the direction of the object's movement.
A cropped image $I_{\text{out}}'$ together with the pre-defined prompt \textit{“Analyze the object’s direction. Choose from: left, right, forward, backward”} are provided as inputs to the VLM.
The output only passes if the output of VLM matches the intended direction specified by the prompt $t$.
\section{MovSafeBench}\label{sec:MovSafeBench}




Leveraging our HazardForge pipeline, we develop MovSafeBench, an MCQ-style benchmark created to assess safety-conscious decision-making in autonomous mobile agents, encompassing both self-driving vehicles and mobile robots.
MovSafeBench comprises four scenario categories and encompasses 13 object types, covering both common and anomalous cases.

\subsection{Benchmark Overview}
\begin{table*}[ht]
\centering
\setlength{\tabcolsep}{6pt}

\caption{MovSafeBench statistics for each scenario and object category.}
\label{tbl:dataset-statistics}
\resizebox{\textwidth}{!}{
\begin{tabular}{l|rrrr|rrrrrrrrr|r}
\toprule
Scenario
 & Human & Motorcycle & Bicycle & Cone & Rocks & Debris & Roadkill & Dog & Cat & Deer & Fox & Pig & Raccoon & Total\\
\midrule
Static & 144 & 331 & 27 & 350 & 348 & 301 & 258 & 289 & 102 & 168 & 123 & 303 & 349 & 3093\\
Motion & 99 & 115 & 123 & - & - & - & - & 109 & 103 & 112 & 113 & 103 & 104 & 981\\
Intrusion & 162 & 356 & 319 & - & - & - & - & 236 & 127 & 130 & 142 & 227 & 300 & 1999\\
Distance & 38 & 198 & 23 & 239 & 169 & 133 & 43 & 95 & 16 & 33 & 19 & 93 & 82 & 1181\\
\bottomrule
\end{tabular}
}
\end{table*} 
This subsection provides an overview of our MovSafeBench.
We select DriveBench~\cite{xie2025DriveBenchvlmsreadyautonomousdriving} for scenes of autonomous driving and SA-Bench~\cite{savlm_icra2025} for scenes of mobile robots to be transformed using HazardForge.
We use 200 images from DriveBench and 200 images from SA-Bench as input for HazardForge.
We remark that the steering decision answer of the benchmark is valid since the answer of the MCQ is attributed to the actual steering in the original dataset.
The detailed distribution of MovSafeBench is shown in Table~\ref{tbl:dataset-statistics}.

\subsection{Evaluation}
Real-world environments are characterized by the presence of both common and anomalous objects subject to complex spatio-temporal dynamics. 
We conduct a systematic evaluation on VLMs to quantify the influence of object typology and spatio-temporal conditions on model performance.
To additionally confirm that the generated object truly influences the scenario and to examine the performance disparity between humans and VLMs, we perform a human evaluation.

\subsubsection{Experimental setting}
We conduct an evaluation of 7 VLMs on MovSafeBench, covering Qwen2.5-VL~\cite{bai2025qwen25vltechnicalreport}, LLaVA-NEXT~\cite{NEURIPS2023_llava}, Phi4-Multimodal~\cite{microsoft2025phi4minitechnicalreportcompact}, InternVL3.5~\cite{wang2025internvl35advancingopensourcemultimodal}, and PaliGemma2~\cite{steiner2024paligemma2familyversatile}.
We additionally assess the initial set of 400 original images from DriveBench and SA-Bench with the same QA pairs to demonstrate the impact of the objects generated by HazardForge.
In the following, we refer to the initial collection of 400 original images as the \textit{no edit} category.
We use 4 NVIDIA H100 SXM5 94GB for each evaluation.

\noindent\textbf{User Study} We randomly sample a subset from the dataset and use it to measure human performance.
We generate three spreadsheets, each containing 300 images, and each spreadsheet is evaluated by a different annotator.
Each spreadsheet contains 60 unique images for every category: original images, \textit{static scenario}, \textit{motion scenario}, \textit{intrusion scenario}, and \textit{distance scenario}.

\subsubsection{Results}
\begin{table}[htbp]
\centering
\scalebox{}{}
\setlength{\tabcolsep}{22pt}
\caption{Scenario-wise comparison. Cyan- and orange-colored cells are the easiest and hardest scenarios for each model.}\label{tbl:scenario-wise}
\resizebox{\columnwidth}{!}{
\begin{tabular}{@{}l|c|c|cccc|c@{}}
\toprule
\multicolumn{1}{c|}{Models} & No edit & Static & Motion & Intrusion & Distance & Total\\
\midrule
Human eval. & 61.1 & 86.1 & 87.8 & 77.8 & 72.8 & 77.1\\
\midrule
InternVL3.5 & 54.8 & \cellcolor{cyan!15}44.9 & \cellcolor{orange!15}25.5 & 42.5 & 39.5 & 40.7\\
$\text{LLAVA-NEXT}_{\text{7B}}$ & 50.3 & 37.2 & \cellcolor{orange!15}32.6 & \cellcolor{cyan!15}37.7 & 35.8 & 36.5\\
$\text{LLAVA-NEXT}_{\text{13B}}$ & 63.5 & \cellcolor{cyan!15}62.1 & \cellcolor{orange!15}2.0 & 58.5 & 48.6 & 50.8\\
Paligemma2 & 46.0 & 28.9 & \cellcolor{orange!15}10.6 & \cellcolor{cyan!15}29.6 & 24.6 & 25.9\\
Phi4 & 60.2 & \cellcolor{cyan!15}53.8 & \cellcolor{orange!15}14.1 & 47.7 & 38.5 & 44.3\\
$\text{Qwen2.5-VL}_\text{7B}$ & 53.0 & 38.8 & \cellcolor{cyan!15}55.8 & 43.8 & \cellcolor{orange!15}32.9 & 41.5\\
$\text{Qwen2.5-VL}_\text{32B}$ & 61.1 & 49.8 & \cellcolor{orange!15}33.1 & \cellcolor{cyan!15}64.8 & 41.5 & 50.3\\
\midrule
Average & 55.6 & 45.1 & \cellcolor{orange!15}24.8 & \cellcolor{cyan!15}46.4 & 37.3 & 41.4\\
\bottomrule
\end{tabular}
}
\end{table} 
\begin{figure}[htp]
    \centering
    \includegraphics[width=1.0\linewidth]{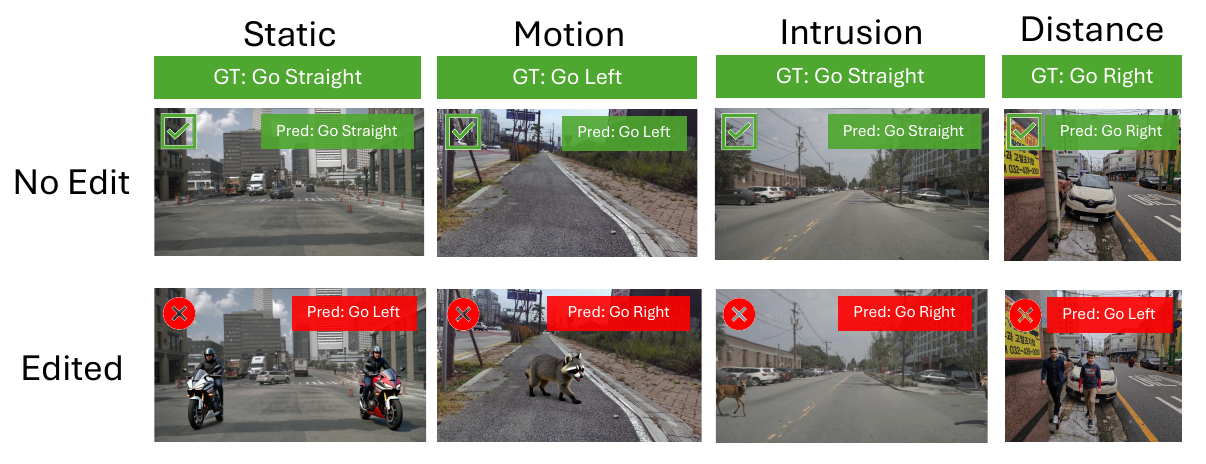}
    \caption{Qualitative results of Qwen2.5-VL-7B in the non-edited scenes and edited scenes.}
    \label{fig:qualitative_nonedit}
\end{figure}
Table~\ref{tbl:scenario-wise} indicates that VLMs underperform when handling temporal dynamics and distant objects. 
\textit{Motion scenario} is the most difficult category for VLMs, which shows the lowest average score 24.8\%, followed by 37.3\% of the \textit{distance scenario} category.
Conversely, the VLMs perform best on \textit{intrusion scenario} (46.4\%) and \textit{static scenario} (45.1\%).
The result suggests that VLMs struggle to infer the motion direction of dynamic objects from images and to determine the appropriate safe maneuver accordingly.
Objects located at a greater distance can be misclassified as dangerous, as suggested by lower scores in the condition of \textit{distance scenario}, which connects to the danger action decision.
In contrast, \textit{static scenario} and \textit{intrusion scenario} are easy for VLMs.

The results in Table~\ref{tbl:scenario-wise} also indicate that introducing additional objects negatively affects the performance of VLMs.
Table~\ref{tbl:scenario-wise} shows the average score of \textit{no edit} as 55.6 \%, which is higher than the other scenarios.
Fig.~\ref{fig:qualitative_nonedit} illustrates how the model’s predictions vary across different scenarios.
The VLM’s predictions can vary when the objects are generated.

As for the comparison of human evaluation and VLM evaluation in Table~\ref{tbl:scenario-wise}, there is a significant gap between human and VLMs for safe decision-making.
In the \textit{static scenario}, the human performance score increases relative to the no-edit condition because the image provides cues indicating that all options other than the ground-truth one are already occupied, whereas the VLM's accuracy decreases compared to the no-edit condition.
Although the recognition of moving objects is easy for humans from a single image given by 87.8\% accuracy, experimental results show that this is not the case for VLMs in particular.

\begin{table*}[ht]
\centering
\setlength{\tabcolsep}{4pt}
\caption{Accuracy ($\uparrow$) of VLMs in the safe path MCQ task across object categories. Orange-colored cells are scores below the total score.}
\label{tbl:category-wise}
\resizebox{\textwidth}{!}{
\begin{tabular}{lc|c|cccc|ccccccccc|ccc}
\toprule
\multicolumn{1}{c}{\multirow{2}{*}{Model}} & \multirow{2}{*}{Size} & No & \multicolumn{4}{c|}{Normal objects} & \multicolumn{9}{c|}{Anomalous objects} & \multicolumn{3}{c}{Total} \\
 &  & edit & human & motorcycle & bicycle & cone & rocks & debris & roadkill & dog & cat & deer & fox & pig & raccoon & Norm. & Anom. & All \\
\midrule
InternVL3.5 & 8B & 54.8 & 43.1 & \cellcolor{orange!20}35.5 & \cellcolor{orange!20}36.4 & 50.8 & 51.1 & 48.4 & 49.2 & \cellcolor{orange!20}38.7 & \cellcolor{orange!20}35.6 & \cellcolor{orange!20}37.7 & \cellcolor{orange!20}37.8 & \cellcolor{orange!20}36.8 & \cellcolor{orange!20}38.1 & \cellcolor{orange!20}40.6 & 40.8 & 40.7 \\
LLAVA-NEXT & 7B & 50.3 & \cellcolor{orange!20}36.3 & \cellcolor{orange!20}33.4 & 38.2 & 40.2 & 38.5 & 37.8 & \cellcolor{orange!20}35.2 & \cellcolor{orange!20}35.1 & \cellcolor{orange!20}36.2 & 39.1 & 39.3 & 36.6 & \cellcolor{orange!20}33.5 & 36.5 & 36.5 & 36.5 \\
LLAVA-NEXT & 13B & 63.5 & \cellcolor{orange!20}43.1 & 56.3 & \cellcolor{orange!20}44.1 & 58.9 & 59.4 & 65.9 & 58.5 & \cellcolor{orange!20}49.5 & \cellcolor{orange!20}31.9 & \cellcolor{orange!20}40.9 & \cellcolor{orange!20}34.3 & \cellcolor{orange!20}50.3 & 53.2 & 52.2 & \cellcolor{orange!20}50.0 & 50.8 \\
Paligemma2 & 10B & 46.0 & 34.3 & \cellcolor{orange!20}23.7 & 28.1 & 27.8 & 27.1 & 31.6 & 27.6 & 33.7 & \cellcolor{orange!20}18.7 & \cellcolor{orange!20}24.2 & \cellcolor{orange!20}22.7 & \cellcolor{orange!20}15.2 & \cellcolor{orange!20}25.3 & 27.4 & \cellcolor{orange!20}25.1 & 25.9 \\
Phi4 & 6B & 60.2 & \cellcolor{orange!20}37.0 & 50.6 & \cellcolor{orange!20}37.0 & 54.5 & 53.2 & 56.5 & 49.2 & \cellcolor{orange!20}42.3 & \cellcolor{orange!20}31.0 & \cellcolor{orange!20}36.3 & \cellcolor{orange!20}32.5 & \cellcolor{orange!20}42.2 & \cellcolor{orange!20}42.9 & 46.5 & \cellcolor{orange!20}43.1 & 44.3 \\
Qwen2.5-VL & 7B & 53.0 & 51.5 & 44.6 & 52.4 & \cellcolor{orange!20}36.2 & 51.5 & \cellcolor{orange!20}41.5 & \cellcolor{orange!20}28.6 & 45.8 & 45.1 & \cellcolor{orange!20}34.8 & 42.8 & \cellcolor{orange!20}33.1 & \cellcolor{orange!20}33.5 & 45.4 & \cellcolor{orange!20}39.5 & 41.5 \\
Qwen2.5-VL & 32B & 61.1 & 55.1 & 56.0 & 53.1 & 55.7 & 53.6 & \cellcolor{orange!20}47.7 & \cellcolor{orange!20}43.5 & 51.0 & \cellcolor{orange!20}44.8 & \cellcolor{orange!20}37.9 & \cellcolor{orange!20}46.4 & \cellcolor{orange!20}45.5 & 52.0 & 55.2 & \cellcolor{orange!20}47.8 & 50.3 \\
\midrule
Average &  & 55.6 & 42.9 & 42.9 & \cellcolor{orange!20}41.3 & 46.3 & 47.8 & 47.1 & 41.7 & 42.3 & \cellcolor{orange!20}34.8 & \cellcolor{orange!20}35.8 & \cellcolor{orange!20}36.5 & \cellcolor{orange!20}37.1 & \cellcolor{orange!20}39.8 & 43.4 & \cellcolor{orange!20}40.4 & 41.4\\
\bottomrule
\end{tabular}
}
\end{table*} 
\begin{figure}[ht]
    \centering
    \includegraphics[width=1.0\linewidth]{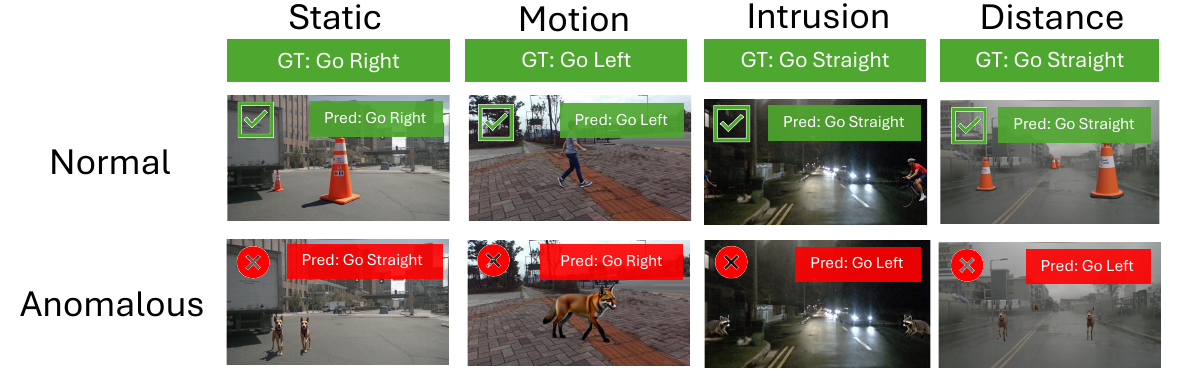}
    \caption{Qualitative comparison of Qwen2.5-VL-7B between normal objects and anomalous objects.}
    \label{fig:qualitative_normanorm}
\end{figure}

Table~\ref{tbl:category-wise} reports the drop in performance for scenes containing anomalous objects, highlighting the difficulty VLMs have in handling such edge cases.
The table presents a category-wise assessment of VLMs for object types such as \textit{human}, \textit{motorcycle}, and \textit{cone}, along with less common or atypical objects in cityscape scenes, like \textit{dog} and \textit{cat}.
\textit{human} and \textit{cone} categories frequently seen in general datasets show 42.9\% and 46.3\%, respectively.
By contrast, \textit{dog} shows 34.8\%, which is the lowest in all categories.
Fig.~\ref{fig:qualitative_normanorm} reveal that normal and anomalous performance changes. 
This finding shows that VLMs perform worse on anomalous objects than on objects that appear frequently.

\begin{table}[htbp]
\centering
\setlength{\tabcolsep}{10pt}
\caption{Benchmark-wise analysis of model performance across the original datasets.}\label{tbl:benchmark-wise}
\resizebox{\columnwidth}{!}{
\begin{tabular}{l|ccccccc|c}
\toprule
Benchmark & InternVL3.5 & LLaVANX-7B & LLaVANX-13B & PaliG2 & Phi-4 & Qwen2.5-VL-7B & Qwen2.5-VL-32B & Avg\\
\midrule
DriveBench & 51.0 & 45.1 & 41.2 & 25.0 & 34.4 & 45.0 & 42.1 & 40.5\\
SA-Bench   & 24.7 & 23.0 & 65.7 & 27.3 & 59.6 & 36.1 & 63.2 & 42.8\\
Overall    & 40.7 & 36.5 & 50.8 & 25.9 & 44.3 & 41.5 & 50.3 & 41.4\\
\bottomrule
\end{tabular}
}
\end{table}
Table~\ref{tbl:benchmark-wise} presents the results for the original dataset, revealing the reduced accuracy of the VLMs across driving scenarios.
Overall, VLM can decide the safe action in the scenes of mobile robots.
By contrast, the accuracy in the driving scene is more degraded than that one in the scene of mobile robots.

\section{Conclusion}\label{sec:conclusion}
This study tackles the problem of insufficient critical scenarios, particularly those involving not only normal but also anomalous objects with various spatio-temporal conditions, for evaluating the decision-making of safe actions on VLMs in mobile agents.
To address the problem, we introduce HazardForge, a benchmark generation framework, and construct the MovSafeBench dataset, which covers both common and anomalous hazardous scenarios with diverse spatio-temporal situations. 
Our experimental findings indicate that current VLMs perform worse in scenarios involving anomalous objects and moving objects.
{
    \small
    \bibliographystyle{abbrv}
    \bibliography{main}
}


\end{document}


\title{Supplementary Material}
%
%
%
%

%

%
%
\clearpage
\setcounter{page}{1}
\section{Appendix}~\label{appendix}
\subsubsection{Front view generation}
Except for \textit{motion image editing}, \textit{intrusion image editing}, and \textit{distance image editing}, we generate the target object in a front-facing view. 
We refer to this generation as \textit{default image editing}. 
Let $t$ be a text prompt that specifies the object orientation as \textit{facing forward}. 
Let $\Omega_k(I_\text{in})$ denote a conditional mask, where $k \in \{\mathrm{L}, \mathrm{C}, \mathrm{R}\}$ corresponds to the left, center, and right regions, respectively. 
Given an input image $I_{\mathrm{in}}$, the front-view generation is defined as
\begin{align*}
    I_{\mathrm{out}} \sim M\!\left(I_{\mathrm{in}},\, \Omega_{k}(I_{\mathrm{in}}),\, t\right),
\end{align*}
where $\Omega_{\mathrm{C}}(I_{\mathrm{in}})$ specifies the region to be edited under the default setting.
In the \textit{static scenario}, \textit{default image editing} is applied to all regions except the ground-truth region $\Omega_{k^*}(I)$, where $k^*$ denotes the index of the ground-truth region.
In the \textit{intrusion scenario}, when the ground-truth region corresponds to $a_{\mathrm{R}}$ or $a_{\mathrm{L}}$, we apply \textit{default image editing} to the center region $\Omega_{\mathrm{C}}(I_{\mathrm{in}})$.
In the \textit{distance scenario}, \textit{default image editing} is applied to all regions except the ground-truth region $\Omega_{k^*}(I_{\mathrm{in}})$.

\section{Limitation}~\label{sec:limitation}
Our approach, while effective in generating diverse hazardous scenarios, has several limitations including simplification of the problem formulation and simplified motion classification. 
First, the evaluation framework is simplified into an MCQ format, which does not fully capture the complexity of real-world decision-making for autonomous agents. 
Second, although dynamic scenarios are included, the representation of object motion is highly simplified and does not accurately reflect real-world diversities. 
Comprehensively covering dynamic scenarios remains for future work.